\documentclass[authoryear]{elsarticle}

\usepackage{todonotes}
\usepackage{tabularx}
\usepackage{amsmath}
\usepackage{booktabs}
\usepackage{caption}
\usepackage[hyphens]{url}
\usepackage{longtable}
\usepackage{subcaption}
\usepackage{siunitx}
\usepackage[margin=3cm]{geometry}
\usepackage{hyperref}
\usepackage{pgfplotstable}
\usepackage{tabularx}
\hypersetup{
	colorlinks=true,
	linkcolor=blue,
	filecolor=blue,
	urlcolor=blue,
	citecolor=blue,
	breaklinks=true
}

\makeatletter
\def\@author#1{\g@addto@macro\elsauthors{\normalsize%
		\def\baselinestretch{1}%
		\upshape\authorsep#1\unskip\textsuperscript{%
			\ifx\@fnmark\@empty\else\unskip\sep\@fnmark\let\sep=,\fi
			\ifx\@corref\@empty\else\unskip\sep\@corref\let\sep=,\fi
		}%
		\def\authorsep{\unskip,\space}%
		\global\let\@fnmark\@empty
		\global\let\@corref\@empty  
		\global\let\sep\@empty}%
	\@eadauthor={#1}
}
\makeatother

\pgfplotsset{compat=1.7}

\begin{document}

\begin{frontmatter}

\title{A Look at the Evaluation Setup of the M5 Forecasting Competition}

\author[it]{Hansika Hewamalage\corref{cor1}}
\author[sydney]{Pablo Montero-Manso}
\author[it]{Christoph Bergmeir}
\author[business]{Rob J Hyndman}
\address{Hansika.Hewamalage@monash.edu, pablo.monteromanso@sydney.edu.au, Christoph.Bergmeir@monash.edu, Rob.Hyndman@monash.edu}
\address[it]{Dept of Data Science and AI, Faculty of IT, Monash University, Australia.}
\address[sydney]{Discipline of Business Analytics, University of Sydney, Australia.}
\address[business]{Dept of Econometrics \& Business Statistics, Business School, Monash University, Australia.}

\cortext[cor1]{Corresponding Author Name: Hansika Hewamalage, Affiliation: Faculty of Information Technology, Monash University, Melbourne, Australia, Postal Address: Faculty of Information Technology, P.O. Box 63 Monash University, Victoria 3800, Australia, E-mail address: Hansika.Hewamalage@monash.edu}

\begin{abstract}
Forecast evaluation plays a key role in how empirical evidence shapes the development of the discipline. Domain experts are interested in error measures relevant for their decision making needs. Such measures may produce unreliable results. Although reliability properties of several metrics have already been discussed, it has hardly been quantified in an objective way. We propose a measure named Rank Stability, which evaluates how much the rankings of an experiment differ in between similar datasets, when the models and errors are constant. We use this to study the evaluation setup of the M5. We find that the evaluation setup of the M5 is less reliable than other measures. The main drivers of instability are hierarchical aggregation and scaling. Price-weighting reduces the stability of all tested error measures. Scale normalization of the M5 error measure results in less stability than other scale-free errors. Hierarchical levels taken separately are less stable with more aggregation, and their combination is even less stable than individual levels. We also show positive tradeoffs of retaining aggregation importance without affecting stability. Aggregation and stability can be linked to the influence of much debated magic numbers. Many of our findings can be applied to general hierarchical forecast benchmarking.

\end{abstract}

\end{frontmatter}

\section{Introduction}
\label{sec:introduction}

The M5 Forecasting Competition has introduced three main innovations compared to the previous versions of the competition: focus on intermittent series, consideration of hierarchies and incorporation of external covariates. Many real-world forecasting problems closely follow the setup of the M5 Competition~\citep[e.g.,][]{bandara2019sales}.

Comparing performance of forecasting models is a difficult task that has received much attention~\citep{FILDES199281, TASHMAN2000437, Armstrong2001}.
If we compare many models on individual time series, the low sample size and temporal dependency tend to produce spurious rankings.
To address this problem, we can average errors across a set of time series to increase the sample size.
However, averaging errors does not always have meaning, e.g., averaging sales and precipitation.

Even when series can be compared,
they may have largely varying scales.
In such scenarios time series with higher scales dominate the overall error,
effectively resulting in the single series scenario mentioned before~\citep{Armstrong2001}.
Scale-invariant error measures tackle some of these issues~\citep{makridakis1993accuracy, hyndman2006evaluation},
they are dimensionless and normalize the scale of the series.
But with scale-free errors we pay the price of losing degrees of importance. For example, if we are forecasting sales of products for budgeting purposes,
it is natural that errors should be proportional to their dollar value.
Hierarchies, by construction, introduce large variation in scale,
so normalizing aggregated time series virtually removes their influence.
Moreover, the way time series are normalized is still an active area of research
with many alternatives~\citep{botchkarev2018performance}.

Consequently, it becomes hard to find an evaluation measure that is both meaningful and does not risk producing spurious results, it seems that there is a trade-off between the two. The dataset used in the experiment is equally important.
There has been much research on the topic~\citep{armstrong1983published}. There is a lack of objective and especially quantitative comparison. Interpretability is of course subjective, but the reliability is not directly compared, only argued in terms such as ``influenced by outliers'', or ``problems with values close to zero'' or ``not a representative sample''. We therefore lack a systematic way of comparing benchmarking methods. For example: How do current evaluation methods fare in this spectrum?
What are the effects of scale-normalization, and how do the scale-free alternatives compare?
Is there a possibility of good trade-offs? Is there room for improvement?
Does the problem fade with larger dataset sizes?
What is the effect of non-independent data?
What is the effect of aggregation? The M5 is an excellent opportunity to study this problem, because it introduced hierarchies and scales (in the form of dollar value of each sales time series) in a domain where errors among series can be compared, featured tens of thousands of time series and hundreds of forecasting methods, all of this in a competition setting where data leakage was almost perfectly controlled.

We study the reliability or spuriousness of an experiment by capturing it in a single number. The number is ``Rank Stability", or how much the ranking produced by an experiment changes when we apply the same setup on ``equivalent'' datasets.  Equivalent datasets are the scenarios where we expect the results of the original experiment to hold, e.g., data of similar nature, or the same data at other time frames. The basis is that if the ranking changes, then the experimental setup is not stable.

In Section \ref{sec:methodology} we define and motivate Rank Stability.
In Section \ref{sec:experimental_setup} a range of experiments are performed to measure the stability of the M5 setup across the two dimensions where the M5 would be extrapolative, similar datasets and other time frames. We compare the M5 setup to alternative popular error measures on the same data, isolating possible causes of instability.
Section \ref{sec:stabexp_splits} shows the stability on similar datasets, approximated by splitting the M5 dataset. The influence of the cross-sectional hierarchical aggregation is finely studied in Section \ref{sec:cross_perlevel_results}, and total aggregation level, collapsing all series and all time-points to one point is in Section \ref{sec:total_agg_exp}. Section \ref{sec:timeframe_exp} is devoted to the stability across time and how it is influenced by aggregation. Section \ref{sec:support_exp} briefly reports the results of the supporting experiments, the agreement in rankings among error measures and how a benchmark procedure might be affected by magic numbers.
In Section \ref{sec:aggregation_tradeoffs}, we show how the tradeoff between stability and the importance assigned to aggregation can be controlled. Finally, Section \ref{sec:conclusions} concludes the paper by summarising the overall findings.

\section{Methodology}
\label{sec:methodology}

%

When benchmarking models using an empirical experiment, 
what we usually want is to obtain the relative performance among the methods, so that the best ones are identified. We are often not interested in the numeric values of the errors of each method.
Also, often the goal is to identify a set of best-performing models, not just one. The methodologies of these top methods are often carefully studied afterwards, featured in the conclusions of the experiment and built upon in further research. They can also be combined in an ensemble for better accuracy.
For these reasons, we argue that to measure the quality of an experiment, the primary object of interest is the ranking it produces.

Moreover, often the resulting ranking in the specific dataset that
was used in the experiment is not the interest. In forecasting, by the time we can measure the errors, the events of interest have already passed and \textit{a posteriori} identifying which were the best methods is rarely useful. We want the ranking obtained in the experiment to be predictive in future and similar applications.
This is the concept behind rank stability, to measure how similar are the rankings produced by a benchmark experiment when applied again in similar circumstances. Designing experiments with good rank stability has been intuitively behind much of the efforts in forecasting research. For example, the use of scale-free error measures to prevent a few series from dominating the error average mitigates a possible source of rank instability, because extrapolating which methods are best from a few time series is not very reliable. In this section, we define a measure of Rank Similarity and Rank Stability to measure the object of interest, the rankings.


\subsection{Benchmarking procedure}
\label{sec:bench_proc}
A benchmarking procedure, setup or experiment is a tuple of a dataset, a set of methods and an error measure.

\subsection{Rank Similarity}
\label{sec:ranksimil}

Comparing rankings has received much interest, because it is an important part of learning-to-rank \citep{burges2005learning} machine learning tasks.
There are many alternatives for comparing rankings \citep{pinto2005weighted}. For simplicity and familiarity,
we choose the most classic, the Spearman’s rank correlation, which is the linear correlation among two ranking vectors.
Given two ranking vectors $R_1$ and $R_2$, the ranking similarity $S$ can be formulated as shown in Equation \ref{eqn:rank_similarity}, where $Cov(R_1, R_2)$ refers to the covariance between $R_1$ and $R_2$ and $\sigma_{R_1}$ and $\sigma_{R_2}$ represent the standard deviations of the $R_1$ and $R_2$ rankings, respectively.

\begin{equation}
\label{eqn:rank_similarity}
	S = \frac{Cov(R_1, R_2)}{\sigma_{R_1}\sigma_{R_2}}
\end{equation}

While no rank similarity measure is perfect, we highlight
two potential issues with Spearman's correlation.
\begin{enumerate}
\item The correlation depends on the number of methods that are being ranked. For example,
the correlation if we flip 1st and 2nd position when ranking 5 methods is not the same as when are ranking 50.
\item It does not matter where in the rankings the differences appear. The correlation if we flip 1st with 2nd or the 49th with 50th is the same.
\end{enumerate}

To reduce the effects from these problems, we consider the correlations at increasingly larger subsets of the methods available from the M5 submissions. For the main experiment, we show the correlations of the Top 10, Top 20, etc. methods, ranked
according to the original results of the M5. This fix is not perfect since we are subsetting models a posteriori and some methods might be left out when they could have performed in the top.

\subsection{Rank Stability}
\label{sec:rankstab}

Rank stability is a measure of the extrapolation capacity of a benchmarking experiment to an equivalent dataset.
Equivalent can mean datasets of similar nature (e.g. in the M5, sales of another set of stores), same dataset in other time periods, etc. It can in general be any scenario for which the original experimental results can be considered relevant or extrapolated to.
We define the rank stability of an experiment as how much the ranking changes on average, across equivalent datasets.
For this, we use the Rank Similarity defined before. Thus, we define the rank stability as \textit{the average rank similarity produced by a benchmark procedure on equivalent datasets.}

Rank stability represents a desirable property of any benchmarking procedure, but it is a necessary,
not sufficient condition. We can have a perfectly stable benchmarking procedure that is useless because it does not rank according to anything of interest (e.g. ranking acording to the alphabetical order of the models' names). 
Rank Stability is not ideal to discern what causes the instability because it captures all potential sources of instability such as model similarity, time series similarity, ill-suited error measures, randomness in the data, etc. and combinations of them. Additionally, Rank Stability captures effects that might be difficult to model explicitly, such as statistical dependency in time series within a dataset, either natural dependence (e.g. similar products in the same store) or because they are hierarchically related.
Similar ideas to Rank Stability have been used to compare search engines \citep{lempel2005rank} (identically named rank stability), or the change across time of the ranking of individuals in a group according to a personality type (named rank-order consistency) \cite{roberts2000rank}.

%

%

\subsection{Different Error Measures}
\label{sec:error_measures}

We calculate rank stabilities for several popular error measures. This is not an exhaustive list of errors, but they comprise variations along several axis: scale-free vs scaled, absolute vs squared error, scaling based on training vs test periods. Therefore, using them in this study allows us to understand how they compare against the M5's WRMSSE measure and pinpoint the potential causes of problems. In Equations \ref{eqn:mae} to \ref{eqn:wape}, $Y_t$ and $F_t$ denote the actual and forecast values at time step $t$, respectively, where $h$ is the length of the forecast horizon and $n$ is the length of the observed period of the series. These definitions are per each series in the dataset, where as for summarising across series, we consider the mean of the per-series errors.

\begin{itemize}
	\item Mean Absolute Error (MAE)
	\begin{equation}
	\label{eqn:mae}
		\textit{MAE} = \frac{1}{h}\sum_{t=n+1}^{n+h}(Y_t - F_t)
	\end{equation}
	\item Symmetric Mean Absolute Percentage Error (SMAPE)
	\begin{equation}
		\label{eqn:smape}
	\textit{SMAPE} = \frac{200}{h}\sum_{t=n+1}^{n+h}(\frac{|Y_t - F_t|}{Y_t + F_t})
	\end{equation}
	\item Mean Absolute Scaled Error (MASE)
	\begin{align}
	\label{eqn:mase}
	\begin{split}
	|q_t| = \frac{Y_t - F_t}{\frac{1}{n-1}\sum_{i=2}^{n}|Y_i - Y_{i-1}|}\\
	\textit{MASE} = \frac{1}{h}\sum_{t=n+1}^{n+h}(|q_t|)
	\end{split}
	\end{align}
	\item Root Mean Squared Scaled Error (RMSSE)
	\begin{align}
	\label{eqn:rmsse}
	\begin{split}
	|sq_t| = \frac{(Y_t - F_t)^2}{\frac{1}{n-1}\sum_{i=2}^{n}|(Y_i - Y_{i-1})^2|}\\
	\textit{RMSSE} = \sqrt{ \frac{1}{h}\sum_{t=n+1}^{n+h}(|sq_t|)}
	\end{split}
	\end{align}

	\item Weighted Absolute Percentage Error (WAPE)
	\begin{equation}
		\label{eqn:wape}
		\textit{WAPE} = \frac{\sum_{t=n+1}^{n+h}|Y_t - F_t|}{\sum_{t=n+1}^{n+h}|Y_t|}
	\end{equation}
\end{itemize}


We introduce the price weighted variants as was done in the M5 Forecasting Competition with the RMSSE to get WRMSSE. For them, we use the prefix PRICE\_, resulting in PRICE\_MAE, PRICE\_SMAPE, PRICE\_MASE, and PRICE\_RMSSE. The PRICE\_RMSSE is the same as the WRMSSE used in the M5 Competition. The price weighting for the base errors is done as in Equation \ref{eqn:price_weighted_error}, where $w_i$ and $e_i$ indicate the weight and the error produced by a particular base error measure for the $i$th time series in the hierarchy. The weight $w_i$ for the $i$th time series is calculated as per Equation \ref{eqn:price_weight}, where $k$ denotes the total number of levels (12 in this case), $\$Sales_i$ is the dollar sales of the $i$th series and $\$Sales_{Total}$ the total dollar sales. The dollar sales in this context refers to the total number of units sold during the last 28 days of the training period multiplied by their respective prices~\citep{makridakism5}.

\begin{equation}
\label{eqn:price_weighted_error}
	\textit{PRICE\_ERROR} = \sum_{i=1}^{42840}w_i * e_i
\end{equation}

\begin{equation}
	\label{eqn:price_weight}
	w_i = \frac{1}{k} * \frac{\$Sales_i}{\$Sales_{Total}}
\end{equation}

\section{Experiments}
\label{sec:experimental_setup}

In this section, we describe the set of experiments carried out on the M5 dataset to analyze the benchmarking procedure using Rank Stability and Rank Similarity.
Rank stability is measured along the two main dimensions of cross-sectionally equivalent datasets, i.e. different datasets but of the same nature, and temporally equivalent, i.e. the same set of time series but for different time frames. We isolate the effect that hierarchical aggregation has on stability, one of the main novelties introduced in the M5. These main experiments are supported by others to check how stability relates to other desirable properties of a benchmark procedure, such as robustness to hyperparameter tuning.

When calculating the rank stability for the scale-free measures WAPE, SMAPE, MASE, and RMSSE, it is calculated by averaging the per-level averages of the errors, so that the series at the top level has around 30k times more weight than a series at the bottom level. The reason for doing this is to exaggerate the influence of scale-free errors in the higher levels so that we can better compare the effect of aggregation on scale-free measures.
This way of computing the scale-free error measures can be formulated as in Equation \ref{eqn:scale_free_error_measure}, where $e_{i,j}$ stands for the error of the $ith$ series at the $jth$ level of the hierarchy, $n_j$ is the total number of series at level $j$ and $k$ is the total number of levels in the hierarchy.

\begin{equation}
	E = \frac{1}{k}\sum_{j=1}^{k}(\frac{1}{n_j}\sum_{i=1}^{n_j}e_{i,j})
	\label{eqn:scale_free_error_measure}
\end{equation}


\subsection{Rank Stability of Cross-sectionally Equivalent Datasets}
\label{sec:stabexp_splits}

Getting access to large datasets of similar nature is generally difficult. In the case of the M5, this is out of our reach.
Therefore, we approximate this ideal experiment by repeatedly splitting a given dataset into two separate datasets having equal number of time series, and comparing the ranking on one set to the other. This way we get separate datasets of similar nature, but with half the sample size of the original dataset. In the case of the M5 Competition, this results in two datasets each having 15245 series at the bottom-most level of the hierarchy. The rank stabilities are averaged over 76 different splits, for computational reasons.

Ideally, the ranking should be very similar in the two halves. The more the ranking on one half of each split deviates from the other, the less rank stability we get. Even though we are greatly reducing the dataset size, halves of each split in M5 dataset already contains around 15000 series (before aggregating). Therefore, if the ranking results between the two halves vary considerably, it is a clear indication of the instability of ranks in the original dataset containing all 30490.

Table~\ref{tab:different_error_measures} presents the rank stability values of the error measures explained in Section \ref{sec:error_measures}. We report these values for the top 5, 10, 20, and 50 methods submitted to the M5 to give a better perspective on how the rank correlation behaves depending on the number of methods. SMAPE and WAPE are the most stable error measures consistenly across all subsets of methods.
The scale-free measures achieve overall higher stability. The only difference between the SMAPE, WAPE, and MASE is how the scale (the denominator) is computed, but MASE has much lower stability.
The RMSSE (partly used in the competition) also has low stability and calculates the scale in a similar way to MASE (an average quantity across the history of the series), pointing towards a possible cause of instability.

We observe in Table~\ref{tab:different_error_measures} that the price-weighted versions of the scale-free errors have considerably reduced stability of their non weighted alternatives. Price-weighting introduces back a form of scaling, undoing the scale normalization and affecting
stability. PRICE\_MAE has a direct and natural interpretation as error in dollars while achieving better stability than the price-weighted scale-free errors. Price-weighted scale-free errors, in this case, combine possible problems coming from the calculation of the scale (values close to zero, abrupt changes during the observed history) with the effects of scaling. In this light, PRICE\_MAE seems a reasonable alternative for both stability and meaningfulness.



\begin{table}[h!]
	\begin{center}
		\begin{filecontents*}{results/different_error_measures.csv}
Error,Top 5,Top 10,Top 20,Top 50
MAE,0.738157894736842,0.688357256778309,0.51203007518797,0.644112562427968
MASE,0.582894736842105,0.496172248803828,0.662010288880095,0.697218037891502
RMSSE,0.576315789473684,0.517703349282297,0.69008705975465,0.702525678872556
SMAPE,0.834210526315789,0.748325358851675,0.827799762564305,0.799833693915927
WAPE,0.834210526315789,0.859170653907496,0.878215275029679,0.875911145035081
PRICE_MAE,0.602631578947368,0.76555023923445,0.582073605065295,0.664438299937319
PRICE_MASE,0.306578947368421,0.374322169059011,0.216481994459834,0.504942424732596
PRICE_RMSSE,0.294736842105263,0.405741626794258,0.242698852394143,0.562671866469863
PRICE_SMAPE,0.621052631578947,0.729186602870813,0.698713889988128,0.61978334715005
\end{filecontents*}
\pgfplotstabletypeset[
		col sep=comma,
		display columns/0/.style={
			column type={l},
			string type,
			postproc cell content/.code={%
				\pgfplotsutilstrreplace{_}{\_}{##1}%
				\pgfkeyslet{/pgfplots/table/@cell content}\pgfplotsretval
			},
		},
		fixed zerofill,
		precision=2,
		every head row/.style={
			before row={\toprule}, 
			after row={\toprule} 
		},
		every last row/.style={after row=\bottomrule}, 
		]{results/different_error_measures.csv} 
	\end{center}
	\caption{Rank Stability of Different Error Measures on Two Subsets of the M5 Competition Dataset}
	\label{tab:different_error_measures}
\end{table}

The quantitative nature of Rank Stability gives us rich information about the benchmarking procedure.
For example, PRICE\_RMSSE (the measure used in the M5), achieves a rank correlation of only 0.29 on the Top 5 subset of methods, on equivalent datasets of more than 15000 time series each. Therefore the PRICE\_RMSSE rankings in this dataset are not random but neither are they particulary extrapolative or reliable.

As mentioned in Section~\ref{sec:experimental_setup}, the scale-free error measures shown in Table~\ref{tab:different_error_measures} are computed by averaging the per-hierarchical-level error averages.
This is not the way scale-free errors are usually reported. In Table~\ref{tab:average_scale_free_error_measures}, we report the rank stabilities of the scale-free errors in the usual way, measured by pooling all the series across all the hierarchical levels and simply taking the average of the errors. This results in perfect stability for the top 5 methods and almost perfect stability for all the other subsets across all the scale-free measures. Thus, when the errors are calculated in this manner, we can get near perfect stability but the influence of the more aggregate levels is now negligible 


\begin{table}[h!]
	\begin{center}
		\begin{filecontents*}{results/average_scale_free_error_measures_all_subsets.csv}
Error,Top 5,Top 10,Top 20,Top 50
MASE,1,0.967942583732057,0.991986545310645,0.993225175404897
RMSSE,1,0.956618819776714,0.987277404036407,0.985794478031421
SMAPE,1,0.988835725677831,0.996497823506134,0.994135056715935
WAPE,1,0.98548644338118,0.982528690146419,0.97822350728916
\end{filecontents*}
\pgfplotstabletypeset[
		col sep=comma,
		display columns/0/.style={
			column type={l},
			string type,
			postproc cell content/.code={%
				\pgfplotsutilstrreplace{_}{\_}{##1}%
				\pgfkeyslet{/pgfplots/table/@cell content}\pgfplotsretval
			},
		},
		fixed zerofill,
		precision=2,
		every head row/.style={
			before row={\toprule}, 
			after row={\toprule} 
		},
		every last row/.style={after row=\bottomrule}, 
		]{results/average_scale_free_error_measures_all_subsets.csv} 
	\end{center}
	\caption{Rank Stability of Scale-Free Error Measures by Pooling All the Series at All Hierarchical Levels Together and Averaging}
	\label{tab:average_scale_free_error_measures}
\end{table}

The results in Tables \ref{tab:different_error_measures} and \ref{tab:average_scale_free_error_measures} show that the difference of the rank stabilities among the different subsets of methods is not very large to affect our qualitative conclusions. Therefore, for simplicity and since more methods allow for better comparison, for the rest of this study we report only the results based on rankings of the top 50 methods.

\subsection{Rank Stability of Cross-sectionally Equivalent Datasets, per Hierarchical Level}
\label{sec:cross_perlevel_results}

\begin{figure*}[htbp!]
	\hspace*{-1.1cm}
	\includegraphics[scale=0.8]{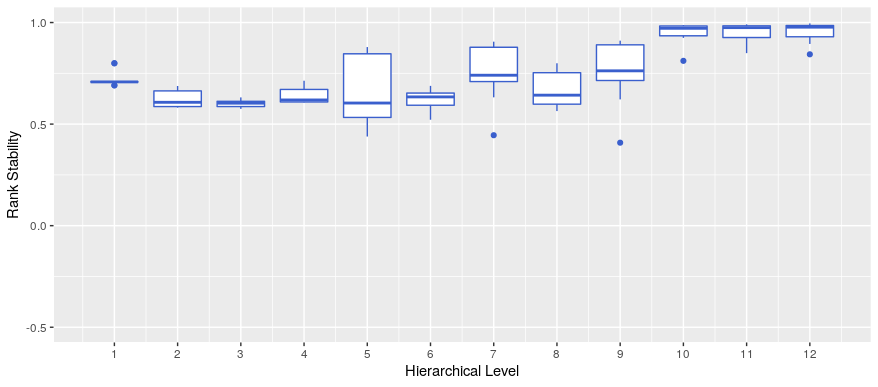}
	\caption{Visualisation of Rank Stability across Different Levels of the Hierarchy in the M5 Dataset}
	\label{fig:per_level_stability}
\end{figure*}

Using the same experiment as in Section \ref{sec:stabexp_splits}, we calculate rank stability at each hierarchical level. The objective is to see the effects of aggregation on rank stability.
We can see from Table~\ref{tab:hierarchical_level_stabilities} and Figure \ref{fig:per_level_stability} that the stability of ranks from all error measures (roughly) decreases the more we aggregate the series.
This highlights the influence of having a few series dominating the error calculation. The highest stability is seen at the bottom-most level (Level 12) for almost all the error measures. The stability of the bottom level is higher than the combination of all the levels, reported in Table~\ref{tab:different_error_measures}, although the combination has many more time series than just the bottom-most level . Having a larger number of time series does not automatically guarantee that the benchmarking will be more stable, it largely depends on the error measure.

\begin{table}[h!]
	\scriptsize
		\begin{filecontents*}{results/hierarchical_level_stabilities.csv}
Level,MAE,MASE,RMSSE,SMAPE,WAPE,PRICE_MAE,PRICE_MASE,PRICE_RMSSE,PRICE_SMAPE
1,0.70783331985361,0.70783331985361,0.799530400145581,0.690775572719736,0.70783331985361,0.70783331985361,0.70783331985361,0.799530400145581,0.690775572719736
2,0.604915633782882,0.581309268657622,0.671961249165942,0.586184716015933,0.586367956002184,0.663371716844936,0.624902693248681,0.687506318620216,0.607315445740745
3,0.602858291040702,0.578028641041713,0.612338748812099,0.604521351881432,0.586379329518572,0.625514335685545,0.599345138200861,0.631431091655377,0.575100592433831
4,0.634858311260287,0.608810431283741,0.705616747881999,0.606720231716441,0.609096032917484,0.670640657540894,0.618776159087692,0.713517550599511,0.612163091170107
5,0.438907788584022,0.846085993893685,0.879390177325758,0.840210940817276,0.860195472834988,0.57935681501102,0.505641264128435,0.603679711668722,0.532735507612674
6,0.592828113310553,0.623031117940838,0.675823189841681,0.688143235537942,0.652836049497543,0.642410326141901,0.566301281921669,0.633973704430111,0.521410012738338
7,0.631726803081465,0.878211122793538,0.883351952200902,0.867973694320319,0.905953656711891,0.445313605758538,0.709428139596013,0.740508168712215,0.733684058879431
8,0.591261095497099,0.740821572274905,0.799452049254908,0.753019036738986,0.757658167701235,0.56425657641992,0.610924641607861,0.642510160341307,0.598220423802495
9,0.62178382231029,0.883257172897669,0.910755808075702,0.890332763815031,0.900658652971268,0.408460885213418,0.714661220858524,0.762301089835615,0.745213013324706
10,0.971552308065592,0.982932143073781,0.986615898659442,0.9873387488121,0.978385263966678,0.811399296358453,0.924401500293184,0.934704642416645,0.968609094769194
11,0.974795023960208,0.991071789635441,0.983785156802879,0.989589441332875,0.977893675313909,0.849504367430293,0.90446498978911,0.92618082374588,0.974534696807328
12,0.977321208322381,0.991852771094082,0.985312999170997,0.99571850294195,0.960311482702145,0.894674919627151,0.843620468285581,0.929889853812403,0.983519774753827
\end{filecontents*}
\pgfplotstabletypeset[
		col sep=comma,
		fixed,
		fixed zerofill,
		precision=2,
		every head row/.style={
			before row={\toprule}, 
			after row={\toprule} 
		},
		display columns/0/.style={
			string type,
		},
		display columns/6/.style={
			column name=PRICE\_MAE,
		},
		display columns/7/.style={
		column name=PRICE\_MASE,
		},
		display columns/8/.style={
		column name=PRICE\_RMSSE,
		},
		display columns/9/.style={
		column name=PRICE\_SMAPE,
		},
		every last row/.style={after row=\bottomrule}, 
		]{results/hierarchical_level_stabilities.csv} 
	\caption{Rank Stability on Different Levels of the Hierarchy in the M5 Dataset}
	\label{tab:hierarchical_level_stabilities}
\end{table}

PRICE\_RMSSE shows a relatively good per-level stability, compared to others. When combining levels (Table \ref{tab:different_error_measures}) the stability greatly decreases, to the point of being considerably worse
than each individual level, suggesting discrepancies in rankings between levels.

\subsection{Rank Stability of Cross-sectionally Equivalent Datasets at Total Aggregation Level}
\label{sec:total_agg_exp}

We calculate the stability for the total aggregation level of the M5 dataset, representing full cross-sectional
and temporal aggregation.
With total aggregation, the sum of all sales of all products through the 28 days of the original test period
results in just a single point.
The purpose of this experiment is to exaggerate the aggregation even further to clarify its effects.
Total aggregation is a reasonable objective in business contexts,
where in addition to measuring the accuracy on a daily basis,
it is also important to achieve good accuracy for the whole 28 days of all products.
For example, a good method on a day-to-day basis is not necessarily good for the sum of the whole period,
and operationally many decisions may be runing at longer time-spans.

The results are as mentioned in Table \ref{tab:total_aggregation_stabilities}.
Total aggregation further decreases the rank stability compared to the cross-sectional top level.
The stability becomes the same for almost all measures because they become mathematically equivalent for ranking,
e.g. RMSE becomes MAE for one data point and scales become irrelevant when comparing on one time series.
Only the SMAPE behaves different.


\begin{table}[h!]
	\begin{center}
		\begin{filecontents*}{results/temporal_aggregation_stabilities.csv}
Error,Total Aggregation,Top Level
MAE,0.656910295812524,0.70783331985361
MASE,0.656910295812524,0.70783331985361
RMSSE,0.656910295812524,0.799530400145581
SMAPE,0.654807913945448,0.690775572719736
WAPE,0.656910295812524,0.70783331985361
PRICE_MAE,0.656910295812524,0.70783331985361
PRICE_MASE,0.656910295812524,0.70783331985361
PRICE_RMSSE,0.656910295812524,0.799530400145581
PRICE_SMAPE,0.654807913945448,0.690775572719736
\end{filecontents*}
\pgfplotstabletypeset[
		col sep=comma,
		fixed,
		fixed zerofill,
		precision=2,
		every head row/.style={
			before row={\toprule}, 
			after row={\toprule} 
		},
		display columns/0/.style={
			column type={l},
			string type,
			postproc cell content/.code={%
				\pgfplotsutilstrreplace{_}{\_}{##1}%
				\pgfkeyslet{/pgfplots/table/@cell content}\pgfplotsretval
			},
		},
		every last row/.style={after row=\bottomrule}, 
		]{results/temporal_aggregation_stabilities.csv} 
	\end{center}
	\caption{Rank Stability Comparison on Total Aggregation Level and Top Level of the Hierarchy in the M5 Dataset}
	\label{tab:total_aggregation_stabilities}
\end{table}

\subsection{Rank Stability across Different Time Frames}
\label{sec:timeframe_exp}

We analyze rank stability among different time frames of the same dataset. We approximate this scenario in the M5 Competition by splitting the 28-days test period of the \textit{full} M5 dataset into two parts, the first 14 and the last 14 days. We check how the ranking changes among the two halves.
This approximation might have the limitation that the horizons are different, so they do not represent exactly comparable datasets since the short/long-term behavior is confounded with the time frame.

According to Table \ref{tab:hierarchical_level_stabilities_rolling_average_aggregated_levels}, all error measures have very small or even negative correlation values. The PRICE\_RMSSE measure shows a correlation of -0.14 between the two time frames, suggesting that the ranking across time is not only unstable, but the better methods the methods are in the first 14 days, the worse they are in the last 14 days.
Rank stability does not tell us about the possible cause. It can be either that the two time frames are ``clearly different'' (structural breaks), that the methods differ greatly in terms of the short range or longer range forecasts (even though the difference is just 14 days) or if the benchmarkings are ``random'' due to problems with the error measures.
If we do not get stable rankings for over 40K time series across consecutive 14 day periods, it is very unlikely that a ranking over 28 days will hold for other time periods. A possible consequence of this result is the need to average over longer time frames to get stable rankings, but this hypothesis should be tested empirically.

\begin{table}[]
	\centering
	\begin{tabular}{lS}
		\toprule
		Error        & {Cor}   \\
		\hline
		MAE          & -0.18 \\
		MASE         & 0.08  \\
		RMSSE        & 0.06  \\
		SMAPE        & 0.10  \\
		WAPE         & -0.07 \\
		PRICE\_MAE   & 0.04  \\
		PRICE\_MASE  & -0.20 \\
		PRICE\_RMSSE & -0.14 \\
		PRICE\_SMAPE & -0.12 \\
		\bottomrule
	\end{tabular}
\caption{Rank Stability on All Levels of the Hierarchy in the M5 Dataset in the Rolling Average Scenario}
\label{tab:hierarchical_level_stabilities_rolling_average_aggregated_levels}
\end{table}

\begin{figure*}[htbp!]
	\hspace*{-1.1cm}
	\includegraphics[scale=0.8]{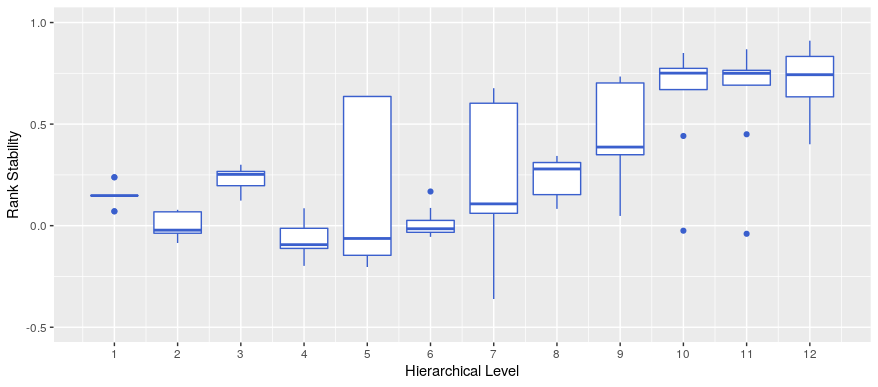}
	\caption{Visualisation of Rank Stability across Different Levels of the Hierarchy in the M5 Dataset, Two Time Frames Scenario}
	\label{fig:per_level_stability_rolling_average}
\end{figure*}

Table \ref{tab:hierarchical_level_stabilities_rolling_average} shows the rank stability results disaggregated by hierarchical level. The same results are illustrated in Figure \ref{fig:per_level_stability_rolling_average} across different levels. The comparison across hierarchical levels sheds some light on the causes of the temporal instability.
At the bottom level, the rankings are more stable than when combining all levels, therefore causes such as difference between short-term and long-term  behavior, structural breaks, etc. lose importance.
Compared to the ``cross-sectional" stability results of Section \ref{sec:cross_perlevel_results}, the temporal case is less stable and degrades much faster when aggregating, though both experiments represent the same amount of data points.

\begin{table}[]
	\centering
	\scriptsize
	\begin{tabular}{cSSSSSSSSS}
		\toprule
		Level & {MAE}   & {MASE}  & {RMSSE} & {SMAPE} & {WAPE}  & {PRICE\_MAE} & {PRICE\_MASE} & {PRICE\_RMSSE} & {PRICE\_SMAPE} \\
		\hline
		1  & 0.15  & 0.15  & 0.24  & 0.07  & 0.15  & 0.15       & 0.15        & 0.24         & 0.07         \\
		2  & -0.03 & 0.06  & 0.07  & -0.02 & 0.08  & -0.05      & -0.04       & 0.07         & -0.09        \\
		3  & 0.27  & 0.26  & 0.12  & 0.24  & 0.29  & 0.30       & 0.25        & 0.16         & 0.20         \\
		4  & -0.01 & -0.10 & -0.07 & -0.12 & -0.09 & 0.09       & -0.11       & 0.00         & -0.20        \\
		5  & -0.20 & 0.64  & 0.64  & 0.60  & 0.64  & -0.06      & -0.15       & -0.13        & -0.17        \\
		6  & 0.09  & -0.02 & -0.03 & 0.01  & -0.02 & 0.17       & -0.05       & 0.03         & -0.06        \\
		7  & 0.01  & 0.60  & 0.65  & 0.60  & 0.68  & -0.36      & 0.06        & 0.11         & 0.09         \\
		8  & 0.15  & 0.28  & 0.34  & 0.31  & 0.31  & 0.34       & 0.14        & 0.18         & 0.08         \\
		9  & 0.21  & 0.68  & 0.70  & 0.72  & 0.73  & 0.05       & 0.35        & 0.39         & 0.35         \\
		10 & 0.44  & 0.67  & 0.75  & 0.85  & -0.02 & 0.73       & 0.75        & 0.77         & 0.82         \\
		11 & 0.45  & 0.69  & 0.75  & 0.71  & -0.04 & 0.80       & 0.76        & 0.75         & 0.87         \\
		12 & 0.56  & 0.91  & 0.74  & 0.83  & 0.40  & 0.84       & 0.63        & 0.74         & 0.79        \\
		\bottomrule
	\end{tabular}
	\caption{Rank Stability on Different Levels of the Hierarchy in the M5 Dataset in the Rolling Average Scenario}
	\label{tab:hierarchical_level_stabilities_rolling_average}
\end{table}


\subsection{Supporting Experiments}
\label{sec:support_exp}

We do additional experiments measuring Rank Similarity when changing the benchmarking procedure itself. Therefore these experiments do not represent the concept of Rank Stability as we define it, which is Rank Similarity across datasets.
This involves magic numbers related experiments as well as comparisons among error measures. 
The idea of magic numbers is to correct the forecasts of a model by multiplying them by a constant to improve accuracy. Magic numbers were used in the M5 Competition, and there was discussion about their methodology and interpretation in the M5 Competition Kaggle forums~\citep{m5magicdiscussion1, m5magicdiscussion2, m5magicdiscussion3} and the ISF conference \cite{prakash2020learningsM5}. The second and the fifth methods in terms of the final rankings of M5 have also used these multipliers~\citep{makridakism5}. 
To illustrate their effect, we experiment by adjusting all methods' forecasts to an optimal magic number. Then we measure how much the ranking of the magically adjusted methods differs from the non-adjusted models, measured in terms of Rank Similarity.
For simplicity and computational efficiency, we compare
three errors, namely PRICE\_RMSSE, MAE and SMAPE at just two hierarchical levels, the top and the bottom. The optimal magic number for each method is found by grid search using 500 equidistanced points in-between 0 and 2. The number that gives the smallest error in the test period is chosen as the magic number for each method, error and hierarchical level. The results from the magic numbers experiment follow stability; the more we aggregate, the higher the effects from magic numbers. For more details on this experiment as well as the rank similarity among error measures experiment, refer to the Appendix.



\section{Aggregation Tradeoffs}
\label{sec:aggregation_tradeoffs}

\begin{figure*}[htbp!]
	\includegraphics[scale=0.5]{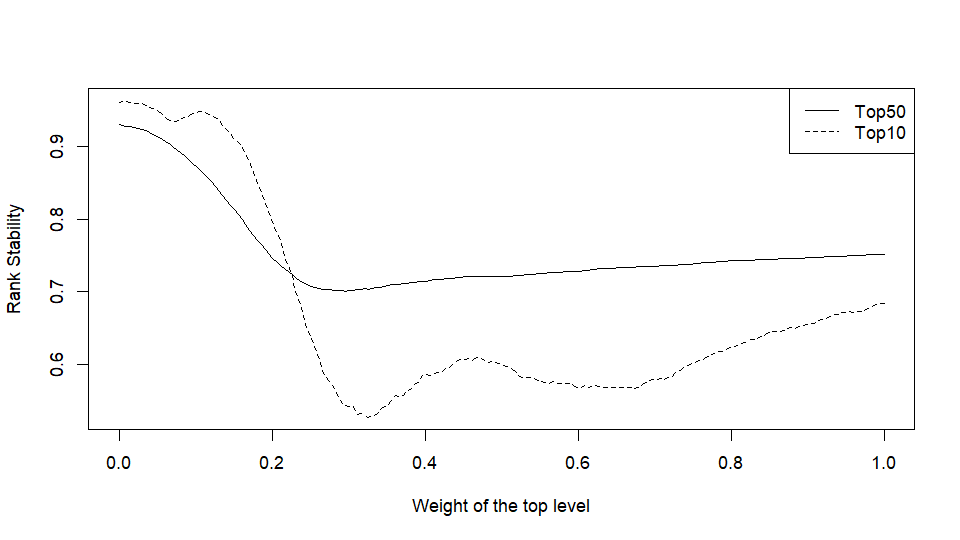}
	\caption{Stability of PRICE\_RMSSE in the M5 Top 50 and Top 10 Methods with Varying Weights for the Top Level}
	\label{fig:top_level_weight_effect}
\end{figure*}

The negative effect of aggregation on stability of benchmarking has been evidenced in our experiments.
When the aggregated time series are given their proportional relevance in an error measure, the rankings become much less stable. On the other hand, the traditional scale-free measures achieve almost perfect stability, but they remove practically all influence of the aggregated levels. But between these two extremes we can find a better tradeoff of practitioners interest and rankings that are reliable/extrapolative. We experiment with controlling the influence of the top level in the PRICE\_RMSSE measure in the experiment of Section \ref{sec:stabexp_splits}, but only considering the bottom and the top level. We introduce a weight for the top level, in the range between 0 and 1. The effect of this weight on the stability is illustrated in Figure \ref{fig:top_level_weight_effect}, for both Top50 and Top10 subsets. The higher the weight of the top level, the lower the rank stability.
However, the stability does not degrade linearly with the weight, which indicates the potential of beneficial tradeoffs.
For small weights of around 0.05 for the Top50 and 0.18 for the Top10, the stability degrades relatively slowly, and it is not much different from the optimal stability achieved with only the bottom level.

A weight of 0.05 might seem small,
but it has the direct interpretation of the top level having a relative importance of 1,500 times an average series in the bottom level (rather than the original 30490). We can still assign considerable importance to the top levels and retain stability. The decrease in stability is non-monotonic, first degrades and the \textit{improves}. We interpret this as the rankings at the bottom level and top level tend to disagree (negative correlation) and therefore their combination reduces stability.
We derive the conclusion that there are non-trivial tradeoffs in stability when considering the importance of higher aggregated series.

\section{Discussion \& Conclusions}
\label{sec:conclusions}

We have found that the ranking produced by the M5 Competition is likely to change if the same setup
is applied to other datasets or even time frames in the same dataset.
The result might seem surprising considering the 42840 time
series that comprise the dataset.
We have formalized the amount of change expected of a benchmarking procedure into a measure that we name Rank Stability.
 We have identified the driving factor of this problem as
 the varying scale or importance among the series in the dataset. In the case of the M5, this was introduced by the error measure and hierarchical aggregation. Long-standing scale-free measures such as MASE
 or SMAPE are much more stable than the M5's WRMSSE (PRICE\_RMSSE in this paper). But, they do not assign proportional importance to the hierarchically aggregated time series.
Introducing importance or scale into the series, either based on price-weighting (as done in the Competition) or proportional to the aggregation level, reduces the stability of all error measures.
The stability is generally lower the more we aggregate, but there are complex interactions between aggregation and error measures. Combining errors of time series across levels results in lower stability than individual levels, but the decrease in stability is relatively stronger in errors such as MASE and RMSSE rather than others such as SMAPE, WAPE or even MAE. The familiy of errors that calculates the scale based on historical data (MASE / RMSSE) has lower stability than the family
that calculates the scale on the test period (SMAPE / WAPE), though this may be particular of the M5 data.

The low rank stability found in the M5 helps explain some phenomena such as the ``Shake-off'' reported in Kaggle, the influence of magic numbers and makes fine-grain conclusions based on the ranking of the M5 less reliable, though conclusions at large, such as the effectiveness of model families like decision trees, are likely to hold.
We have identified the difficulties in benchmarking aggregation, which are likely to be relevant outside of the M5, in general hierarchical forecasting. So far it seems difficult to simultaneously achieve rank stability while considering scale differences, but we have found that there are positive tradeoffs allowing us to get most of the stability while sacrificing a fraction of the scale. Also, some error measures are better than others, for example PRICE\_MAE is more stable than
PRICE\_RMSSE in the M5 and also does not depend on arbitrary normalization.
Rank Stability will help analyize and design experiments in the future, and hopefully the results reported here on the M5 will raise awareness to the nuances of the benchmarking process.

\section*{Acknowledgement}
This research was supported by the Australian Research Council under grant DE190100045, Facebook Statistics for Improving Insights and Decisions research award and Monash University Graduate Research funding.

\bibliographystyle{elsarticle-harv}
\bibliography{references}

\appendix

\section{Supporting Experiments: Magic Numbers}

The magic numbers experiment acts as a proxy for the susceptibility of benchmarking procedures.
A magic number can be seen as one hyperparameter that affects the forecast in a very simple way
(just a product) and yet can have an enormous influence on the ranking.
For example, this hyperparameter might have been adjusted in an experiment where data leakage
is possible or, in a competition, when simply picking among many alternatives that differ
only in this hyperparameter. As an exaggerated example, at the total aggregation level, a properly tuned magic number can give us perfect forecasts for any method. The influence of magic numbers can be a consequence of the lack of degrees of freedom.

We report the results of the experiment in Table \ref{tab:magic_numbers_similarity}.
These results comply with the general stability experiments; stability decreases the more we aggregate.
There is also a strong dependence on the error measure. For example, for the SMAPE at the bottom level, all methods have an optimal magic number of 0, and thus results in no ranking.
In summary, while the effect of this ``magic'' hyperparameter due to aggregation follows stability or is predicted by it (less stable procedures are more susceptible to fine tuning), a procedure such as the SMAPE bottom level can be very stable and still be affected by these problems.

\begin{table*}
	\begin{center}
		\begin{tabular}{lcc}
			\toprule
			Error & Top Level & Bottom Level\\
			\hline
			MAE & 0.55 & 0.91\\
			PRICE\_RMSSE & 0.62 & 0.98\\
			SMAPE & 0.64 & *\\
			\bottomrule
		\end{tabular}
		\caption{Rank Similarity with Magic Numbers at the Top and Bottom Levels of the Hierarchy in the M5 Competition Dataset}
		\label{tab:magic_numbers_similarity}
	\end{center}
\end{table*}

\section{Supporting Experiments: Rank Similarity among Error Measures}
\label{sec:appendix}

We measure the rank similarity of all error measures by using the complete dataset, and the same forecasting methods to compare the rankings produced by other error measures against the original ranking of the M5 Competition. This experiment can give futher perspective on the effects of the error by contrasting their specific differences, e.g. scale-free vs price-weighted.
These results are shown in Table \ref{tab:discrepancy_error_measures}. From this experiment, we can see that the more stable error measures such as SMAPE and WAPE as indentified in Section \ref{sec:stabexp_splits}, are not much correlated to the PRICE\_RMSSE measure. Interestingly, PRICE\_MAE is also not strongly correlated to PRICE\_RMSSE.
Correlations with PRICE\_MAE are low in general.  The largest agreement between price-weighted and nonweighted versions of the scale-free measures is seen on MASE, while the largest disagreement between the price weighted and nonweighted versions is seen on MAE. An important caveat to all these results is that the less stable a measure, the less reliable are these correlations.

%

\begin{table}[h!]
	\footnotesize
	\hspace{-1cm}
	\begin{filecontents*}{results/discrepancy_error_measures.csv}
Error,MAE,MASE,RMSSE,SMAPE,WAPE,PRICE_MAE,PRICE_MASE,PRICE_RMSSE,PRICE_SMAPE
MAE,1,0.68,0.63,0.62,0.66,0.26,0.86,0.79,0.78
MASE,0.68,1,0.94,0.89,0.73,0.27,0.76,0.7,0.57
RMSSE,0.63,0.94,1,0.84,0.7,0.21,0.7,0.73,0.5
SMAPE,0.62,0.89,0.84,1,0.67,0.01,0.6,0.49,0.68
WAPE,0.66,0.73,0.7,0.67,1,0.2,0.53,0.54,0.39
PRICE_MAE,0.26,0.27,0.21,0.01,0.2,1,0.47,0.5,-0.05
PRICE_MASE,0.86,0.76,0.7,0.6,0.53,0.47,1,0.93,0.71
PRICE_RMSSE,0.79,0.7,0.73,0.49,0.54,0.5,0.93,1,0.56
PRICE_SMAPE,0.78,0.57,0.5,0.68,0.39,-0.05,0.71,0.56,1
\end{filecontents*}
\pgfplotstabletypeset[
	col sep=comma,
	display columns/0/.style={
		column type={l},
		string type,
		postproc cell content/.code={%
			\pgfplotsutilstrreplace{_}{\_}{##1}%
			\pgfkeyslet{/pgfplots/table/@cell content}\pgfplotsretval
		},
	},
	fixed zerofill,
	precision=2,
	fixed,
	every head row/.style={
		before row={\toprule}, 
		after row={\toprule} 
	},
	display columns/6/.style={
		column name=PRICE\_MAE,
	},
	display columns/7/.style={
		column name=PRICE\_MASE,
	},
	display columns/8/.style={
		column name=PRICE\_RMSSE,
	},
	display columns/9/.style={
		column name=PRICE\_SMAPE,
	},
	every last row/.style={after row=\bottomrule}, 
	]{results/discrepancy_error_measures.csv} 
	\caption{Rank Similarity Matrix of All Error Measures}
	\label{tab:discrepancy_error_measures}
\end{table}

\end{document}